# LL-GaussianImage: Efficient Image Representation for Zero-shot Low-Light Enhancement with 2D Gaussian Splatting

Yuhan Chen, Wenxuan Yu, Guofa Li, *Senior Member, IEEE,* Yijun Xu, Ying Fang, Yicui Shi, Long Cao, Wenbo Chu, Keqiang Li

*Abstract*—2D Gaussian Splatting (2DGS) is an emerging explicit scene representation method with significant potential for image compression due to high fidelity and high compression ratios. However, existing low-light enhancement algorithms operate predominantly within the pixel domain. Processing 2DGS-compressed images necessitates a cumbersome decompression-enhancement-recompression pipeline, which compromises efficiency and introduces secondary degradation. To address these limitations, we propose LL-GaussianImage, the first zero-shot unsupervised framework designed for low-light enhancement directly within the 2DGS compressed representation domain. Three primary advantages are offered by this framework. First, a semantic-guided Mixture-of-Experts enhancement framework is designed. Dynamic adaptive transformations are applied to the sparse attribute space of 2DGS using rendered images as guidance to enable compression-as-enhancement without full decompression to a pixel grid. Second, a multi-objective collaborative loss function system is established to strictly constrain smoothness and fidelity during enhancement, suppressing artifacts while improving visual quality. Third, a two-stage optimization process is utilized to achieve reconstruction-as-enhancement. The accuracy of the base representation is ensured through single-scale reconstruction and network robustness is enhanced. High-quality enhancement of low-light images is achieved while high compression ratios are maintained. The feasibility and superiority of the paradigm for direct processing within the compressed representation domain are validated through experimental results.

*Index Terms*—Image enhancement, Gaussian Splatting, Image compression.

This work was supported by the National Natural Science Foundation of China under Grant No. 52272421. (*Corresponding author: Guofa Li*).

Yuhan Chen, Wenxuan Yu, Guofa Li, Ying Fang, Yicui Shi and Long Cao are with College of Mechanical and Vehicle Engineering, Chongqing University, Chongqing, 400044, China. (e-mail: 20240701028@stu.cqu.edu.cn; wenxuanyu@cqu.edu.cn; liguofa@cqu.edu.cn; yingfang@stu.cqu.edu.cn; 20212645@cqu.edu.cn; caolong@stu.cqu.edu.cn).

Yijun Xu is with School of Electronic Information, Wuhan University, 430072, China. (e-mail: xyj021021@whu.edu.cn)

Wenbo Chu is with National Innovation Center of Intelligent and Connected Vehicles, Beijing 100089, China (e-mail: chuwenbo@wicv.cn).

Keqiang Li is with School of Vehicle and Mobility, Tsinghua University, Beijing 100084, China (e-mail: likq@tsinghua.edu.cn)

The code is available at https://github.com/YuhanChen2024/LL-GaussianImage.

## I. INTRODUCTION

IMAGE signals are inherently continuous and complex, yet computer vision has long relied on discrete pixel grids as the foundational representation. While this classical format has facilitated the growth of digital image processing, its discrete nature imposes inherent limitations achieving higher fidelity and extreme compression ratios [1-4]. Following revolutionary breakthroughs in 3D scene reconstruction via Neural Radiance Fields (NeRF) and 3D Gaussian Splatting (3DGS), explicit Gaussian representations have emerged as a promising alternative for 2D images [5-6]. As an explicit scene representation, 2D Gaussian Splatting (2DGS) models an image as a collection of anisotropic Gaussians defined by attributes such as position, color, opacity, and covariance. This approach inherits the real-time rendering and high-fidelity capabilities of 3DGS while demonstrating superior potential in image compression compared to conventional pixel-based or Implicit Neural Representation (INR) methods [7-9]. The synergy between compression and representation establishes a robust framework for next-generation image processing technologies.

In practical scenarios such as autonomous driving and security, images frequently suffer from low-light conditions due to insufficient illumination or underexposure. This degradation compromises visual quality and impedes downstream high-level vision tasks, including object detection, semantic segmentation, and cross-view tracking [10-12, 20-21]. To address these challenges, various deep learning-based enhancement methods have been developed. These approaches utilize powerful non-linear fitting capabilities to achieve brightness restoration, color correction, and noise suppression at the pixel level [13-28, 34-38]. However, existing methods operate exclusively in the pixel domain—a dependency that creates significant efficiency and quality bottlenecks when processing 2DGS compressed images. Current pipelines must execute a sequence of decompression, pixel-domain enhancement, and subsequent re-compression. This workflow introduces heavy computational redundancy and cumulative errors, undermining the real-time rendering advantages of 2DGS. Furthermore, noise and artifacts from the enhancement stage are often magnified during re-compression, causing secondary damage to image information [29-31]. Consequently, achieving illumination enhancement directly within the compressed representation domain while maintaining high compression ratios is a critical and challenging problem.

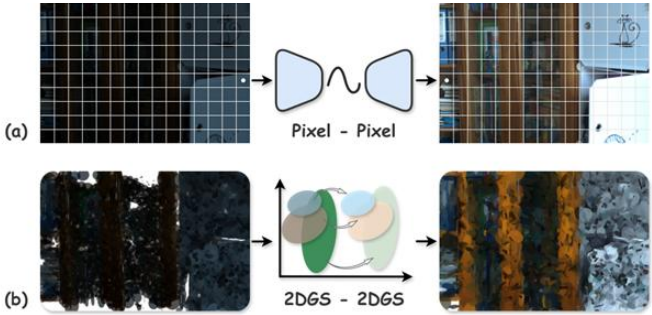

**Fig. 1.** Comparison of low-light image enhancement schemes across (a) pixel domain and (b) 2DGS compressed representation domain. LL-GaussianImage is the first work to perform enhancement directly in the 2DGS compressed domain.

To address these limitations and bridge the gap between explicit compressed representations and low-level vision tasks, this work draws inspiration from recent advancements in 2DGS image compression and Gaussian semantic editing [1-4, 29-33]. It is recognized that illumination attributes reside within both the pixel values and the spatial distribution and attribute parameters of 2D Gaussians. Consequently, LL-GaussianImage is proposed as the first zero-shot unsupervised framework for low-light enhancement directly within the 2DGS compressed domain. This framework introduces a novel compression-as-enhancement paradigm that explicitly decouples geometric fitting from illumination adjustment. The direct enhancement process within the 2DGS compressed representation domain is illustrated in Fig. 1. Specifically, a 2DGS compression strategy first achieves robust reconstruction of image structures to ensure geometric accuracy. Then, geometric attributes such as position and covariance are frozen and a unified Gaussian enhancement network is utilized for the optimization of illumination attributes. Multi-scale spatial features are extracted from rendered views using a lightweight CNN backbone to guide a set of learnable Multi-layer Perceptron (MLP) color operators. By dynamically predicting spatial mixing weights, the model fuses operator outputs in a residual form. This process applies spatially adaptive modulation to Gaussian color attributes, restoring visibility while preserving original data distribution characteristics.

A physics-oriented multi-objective loss function system is designed to constrain the zero-shot training process and prevent color distortion. The objective function incorporates a robust hue preservation term in the HSV space to maintain chromatic fidelity, a histogram contrast term to correct the global illumination distribution, and a spatial consistency constraint to suppress artifacts. These constraints enable the model to learn optimal enhancement parameters without paired ground-truth data. The primary contributions are summarized as follows:

- LL-GaussianImage is proposed as the first zero-shot unsupervised framework for low-light enhancement in the 2DGS compressed domain, pioneering the compression-as-enhancement paradigm. High-fidelity low-light image restoration is achieved by decoupling geometric reconstruction from illumination enhancement. The structural integrity of the compressed representation is strictly maintained throughout the process.

- A semantic-guided Mixture-of-Experts enhancement framework is designed. Multiple residual MLP color operators are orchestrated through a lightweight network. This framework achieves fine-grained and spatially varying adjustments to Gaussian color attributes while balancing local detail enhancement with global illumination restoration.

- A robust zero-shot loss function is constructed through the integration of novel hue preservation constraints and histogram contrast regularization. These constraints guide the optimization process, ensure natural illumination and vivid colors while suppressing secondary artifacts inherent in compressed domain processing.

## II. RELATED WORKS

LL-GaussianImage is situated at the intersection of explicit neural representations and low-level vision tasks. It focuses on the emerging paradigm of direct image enhancement within the compressed domain. Consequently, related works are organized into three categories: recent advancements in low-light image enhancement, an overview of the development of Gaussian Splatting, the overall development of image in domain compression.

### A. Low-light image enhancement

Low-light image enhancement (LLIE) aims to improve the quality of images captured under insufficient illumination by restoring brightness and contrast while suppressing noise. Early traditional techniques primarily relied on histogram equalization and Retinex theory. Specifically, LIME [14] estimates the illumination map by applying structural priors through optimization. However, these handcrafted prior-based methods often depend on complex parameter tuning. Such approaches frequently struggle to balance noise suppression and detail restoration in highly complex scenes.

The field has been increasingly dominated by data-driven methods due to superior performance enabled by advancements in deep learning. Unsupervised learning paradigms have been adopted because paired low-light and normal-light images are difficult to obtain in practical scenarios. EnlightenGAN [19] utilizes generative adversarial networks to enhance images without paired supervision. A series of zero-reference deep curve estimation methods were developed by Guo et al. [15-16] to reduce computational complexity and enhance robustness. These approaches adjust the dynamic range of images by iteratively estimating pixel-level high-order curves to eliminate the need for reference images. Similarly, Pan et al. [23] optimized the curve estimation process via Chebyshev polynomials to achieve superior performance. Regarding the challenge of limited training data, Fu et al. [25] explored learning enhancers from paired low-light instances. Zhang et al. [26] introduced a noise autoregressive paradigm to achieve denoising and illumination enhancement without task-specific data, effectively expanding the boundaries of learning-based methods.

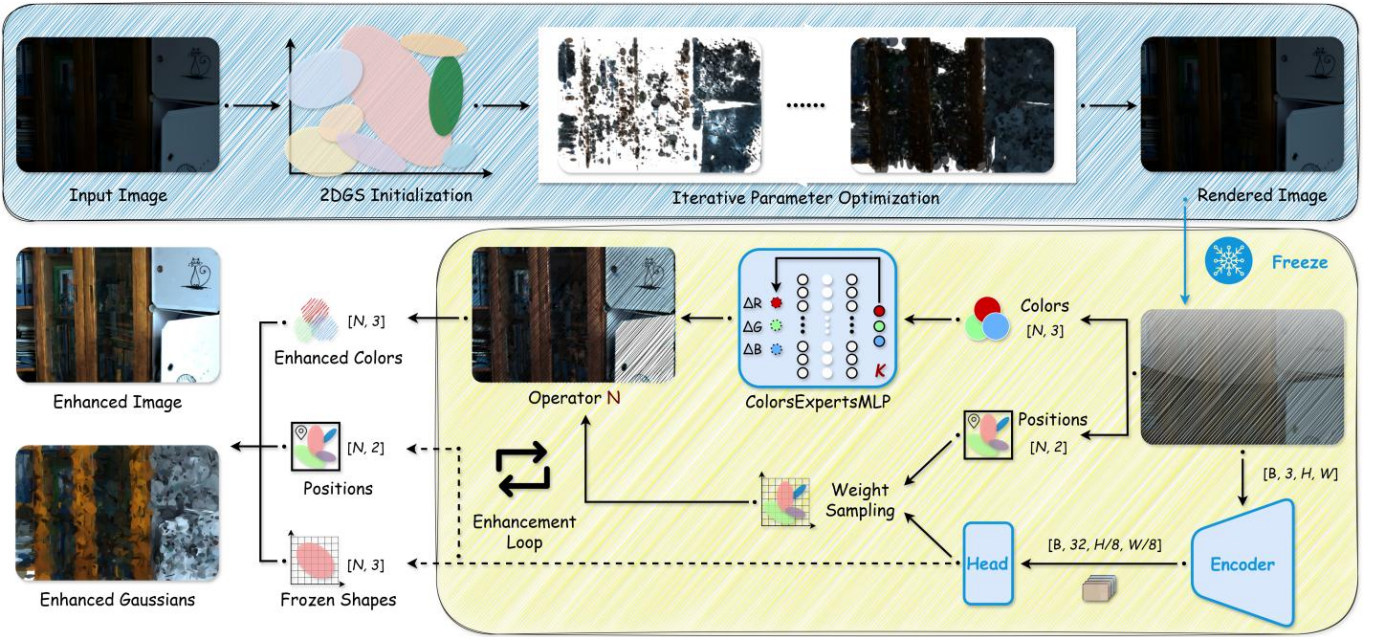

**Fig. 2.** Overview of the proposed two-stage training framework for LL-GaussianImage. In Stage 1, a 2DGS compression-reconstruction strategy is employed for robust structural reconstruction to ensure the geometric accuracy of the explicit representation. In Stage 2, geometric attributes (position and covariance) are frozen, while a lightweight CNN backbone extracts multi-scale spatial features from rendered views to guide a set of learnable MLP color operators. Dynamically predicted spatial mixing weights facilitate the fusion of operator outputs in a residual form. This process modulates Gaussian color attributes in a spatially adaptive manner to restore visibility while preserving the original data distribution.

Despite significant progress, pure data-driven methods are often regarded as black boxes lacking physical interpretability. Several recent approaches address this limitation by integrating physical models into deep architectures. Liu et al. [18] designed a Retinex-inspired unfolding architecture that incorporates a cooperative prior search strategy. Zheng et al. [22] proposed an adaptive unfolding Total Variation network to map the iterative steps of traditional optimization into learnable modules. Furthermore, Wu et al. [27] introduced URetinex-Net, which unfolds the Retinex decomposition into an implicit regularization model to effectively combine data-driven and model-driven methodologies.

Beyond the enhancement of interpretability, lightweight architectures and real-time performance have emerged as significant focal points within the field alongside interpretability due to the proliferation of mobile applications. Ma et al. [17] developed a fast, flexible, and robust illumination enhancement framework. To accommodate resource-constrained devices, Liu et al. [24] introduced EFINet, which improves efficiency through an iterative strategy of enhancement and fusion. Recent architectural designs emphasize efficiency, as demonstrated by the multi-scale residual (FMR-Net) and re-parameterized residual (FRR-Net) networks proposed by Chen et al. [20-21]. Additionally, Bai et al. [13] and Chen et al. [52] explored ultra-lightweight architectures to achieve real-time processing at minimal computational costs.

Generative models and implicit neural representations offer innovative perspectives for low-light illumination enhancement. Specifically, Jiang et al. [34] and Lin et al. [35] explored Retinex-based latent diffusion models and training-free guided diffusion strategies to generate high-quality normal-light images within unsupervised settings. Huang et al. [36] subsequently proposed a zero-shot latent diffusion enhancement method to further improve visual quality. To overcome the limitations of convolutional neural networks regarding resolution and continuity, Yang et al. [37] integrated implicit neural representations into collaborative enhancement tasks. Similarly, Chobola et al. [38] achieved efficient and detail-rich image reconstruction through context-based fast neural implicit representations. The ZERO-IG framework [28] illustrates the potential of zero-shot illumination-guided joint denoising and adaptive enhancement. These advancements suggest that the integration of robust generative priors and innovative feature representations provides a critical pathway for addressing complex low-light degradation.

### B. Gaussian Splatting

3D Gaussian Splatting is established as a fundamental explicit scene representation paradigm in computer graphics and vision [6]. This approach addresses the training and rendering efficiency limitations of traditional implicit Neural Radiance Fields by parameterizing scenes as collections of discrete and learnable 3D Gaussian primitives [6]. Each primitive possesses anisotropic geometric attributes including position, rotation, and scale alongside appearance attributes such as opacity and spherical harmonic coefficients. A customized tile-based differentiable rasterization pipeline enables the synthesis of photo-realistic novel views. This framework achieves real-time or ultra-real-time rendering

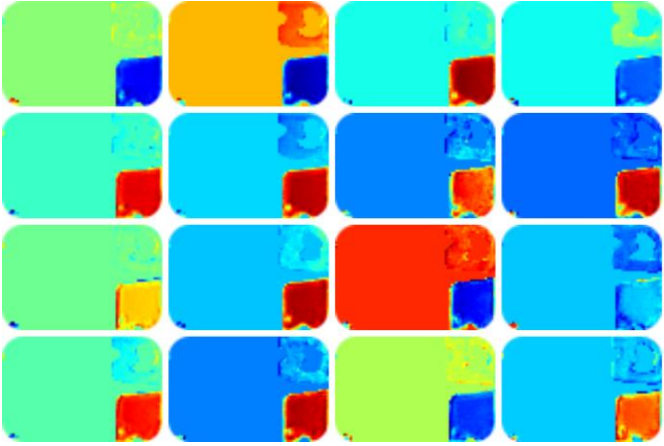

**Fig. 3.** Visualization of semantic mixing weights for $K = 16$. The unsupervised model spontaneously generates attention maps with semantic awareness., where individual weight channels focus on distinct semantic regions to guide spatially adaptive color operations.

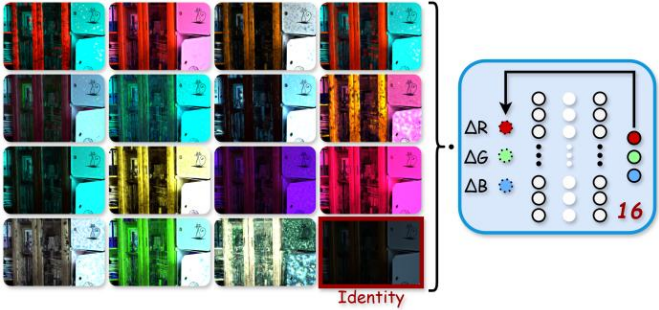

**Fig. 4.** Visual decomposition of the Mixture-of-Experts enhancement module. The global application of each color operator $\Phi_k$ is visualized, where the final identity operator serves as a stable anchor. The system learns a dictionary of diverse exposure and tonal styles, which are then fused locally to generate the final enhanced result.

speeds and provides a robust mathematical foundation for complex scene modeling.

3DGS has been rapidly extended to large-scale dynamic scene modeling and autonomous driving simulation through efficient explicit representation capabilities. Street Gaussians and Periodic Vibration Gaussian capture non-rigid object motions in complex urban environments by introducing temporal dimensions and vibration models [39, 44]. For specific large-span spatiotemporal motion reconstruction, recent methods incorporate physical events to enhance accuracy [42-43]. In the context of autonomous driving perception, DrivingGaussian implements composite Gaussian splatting for surround-view dynamic scenes. DriveDreamer4D and ReconDreamer integrate world models to achieve high-quality 4D driving scene representation and reconstruction [45, 49-50]. To overcome storage and rendering challenges in large-scale scenarios, CityGaussian and Momentum-GS employ tile-based training or momentum distillation strategies to maintain high fidelity while achieving city-level real-time rendering [47-48].

Current advancements emphasize reconstruction efficiency and generative capabilities alongside breakthroughs in scene scale and dynamics. Speedy-Splat [40] increases computational efficiency during training and rendering by utilizing sparse pixels and primitives. Similarly, MVSGaussian [41] and

GaussianPro [46] improve the generalizability and robustness of Gaussian representations via multi-view stereo priors and progressive propagation strategies. Gaussian Splatting also extends into generative tasks. For instance, GaussianDreamer [51] bridges 2D diffusion models with 3D Gaussian representations to facilitate rapid text-to-3D asset generation. These developments establish 3DGS as a flexible and versatile visual primitive. The explicit nature of this representation provides a robust foundation for its adaptation to broader vision tasks.

### C. Gaussian Splatting Based Image Representation

The success of 2DGS in 3D scenes has inspired the adaptation of this paradigm for 2D image representation. The GaussianImage framework was introduced by Zhang et al. as a pioneering effort to adapt 2DGS for compact image representation [3]. This framework discretizes image signals using eight parameters covering position, covariance, and color. An efficient rendering algorithm based on cumulative summation replaces the traditional alpha blending mechanism to improve computational efficiency. This design significantly reduces memory consumption and achieves rendering frame rates far exceeding those of implicit representations. Furthermore, integrating vector quantization techniques facilitates the construction of efficient image codecs. This strategy achieves competitive rate-distortion performance while significantly reducing storage requirements. Based on this framework, Zeng et al. [2] develop a generalizable and adaptive representation mechanism that dynamically allocates Gaussian primitives to improve fitting flexibility. Jiang et al. [4] apply sparse Gaussian representations to dataset distillation and demonstrate the effectiveness of this paradigm for feature extraction. To address large-scale image representation, Zhu et al. [1] introduce multi-level Gaussian Splatting strategies to mitigate detail loss at high resolutions.

As foundational representation capabilities mature, 2DGS-based methods have expanded into complex downstream vision tasks. In super-resolution (SR), GaussianSR and Pixel-to-Gaussian [30-31] overcome traditional grid pixel constraints. These methods achieve arbitrary-scale image reconstruction and detail enhancement by assigning learnable Gaussian kernels to each pixel or establishing continuous pixel-to-Gaussian mappings. Omri et al. [29] further expand these semantic boundaries by utilizing 2DGS as a compressed representation to facilitate efficient vision-language model alignment. Such advancements underscore the potential of this representation for multimodal tasks.

### III. PROPOSED METHOD

The LL-GaussianImage framework is presented in this section. As illustrated in Fig. 2, the architecture is partitioned into two components consisting of 2DGS-based image primitive reconstruction and semantic-guided zero-shot unsupervised Gaussian Splatting enhancement. LL-GaussianImage is established as the initial framework to achieve low-light image enhancement within the 2DGS compressed representation domain. This development establishes a foundation for future low-level vision tasks involving 2DGS-based compressed images. Subsequent sections provide a comprehensive analysis of the methodology and architecture.

### A. 2D Gaussian Splatting-based Image Primitive Reconstruction

The input low-light image $I_{low} \in \Re^{H \times W \times 3}$ is transformed into a set of discrete and differentiable geometric primitive representations in the initial stage of LL-GaussianImage. Distinguishable from conventional pixel grids or implicit neural representations, 2DGS is utilized as the explicit parameterized carrier for the image. This discretized format compactly preserves high-frequency texture details and establishes a geometry-aware data structure for subsequent color-space decoupling and illumination enhancement.

**Explicit Parameterization of 2D Gaussian Primitives.** Specifically, an image is modeled as a set $\Im = \{G_1, G_2, ..., G_N\}$ where $N$ denotes the total number of Gaussian primitives. Each 2DGS primitive $G_i$ is defined as a probability density distribution over the image plane $\Omega \in \mathbb{R}^2$. Unlike 3DGS, these primitives are established directly within the 2D image space to eliminate geometric ambiguities inherent in projection processes. The $i$-th primitive is uniquely determined by a set of learnable parameters:

$$\Theta_i = \mu_i, \Sigma_i, c_i, o_i, \tag{1}$$

where $\mu_i = (u_i, v_i)^T \in \Re^2$ represents the central coordinates of the primitive on the image plane; $c_i \in \Re^3$ denotes the RGB color attributes; $\Sigma_i \in \Re^{2 \times 2}$ is the 2D covariance matrix that dictates the anisotropic shape and spatial extent of the primitive; and $o_i \in [0, 1]$ signifies the opacity, which is utilized to govern the occlusion relationship with the background. Mathematically, the response value of the $i$-th Gaussian primitive at pixel coordinate $p_i \in \Re^2$ is defined by the probability density function as follows:

$$G_i(p) = \exp\left(-\frac{1}{2}(p - \mu_i)^T \Sigma_i^{-1}(p - \mu_i)\right). \tag{2}$$

Ensuring the physical validity and positive semi-definiteness of the covariance matrix $\Sigma_i$ necessitates a specific parameterization during optimization. Rather than performing direct gradient descent on the elements of $\Sigma_i$, the matrix is decomposed into the product of a scaling matrix $S_i$ and a rotation matrix $R_i$:

$$\Sigma_i = R_i S_i S_i^T R_i^T = \begin{bmatrix} cos\theta & -sin\theta \\ sin\theta & cos\theta \end{bmatrix} \begin{bmatrix} s_x & 0 \\ 0 & s_y \end{bmatrix} \\ \begin{bmatrix} s_x & 0 \\ 0 & s_y \end{bmatrix}^T \begin{bmatrix} cos\theta & -sin\theta \\ sin\theta & cos\theta \end{bmatrix}^T . \tag{3}$$

This parameterization facilitates adaptive stretching and rotation of Gaussian primitives to accurately model edges and texture flows. Consequently, these primitives manifest as elongated ellipses in high-frequency regions and isotropic disks in low-frequency areas.

**Accelerated computation via Conic matrix representation.** An implicit conic parameterization method circumvents the frequent calculation of the inverse covariance matrix $\Sigma_i^{-1}$ for individual pixels to improve computational efficiency. The inverse covariance matrix is defined as the Conic matrix $Q_i$ based on the properties of the Gaussian exponential function:

$$Q_i = \Sigma_i^{-1} = \begin{bmatrix} A & B \\ B & C \end{bmatrix}. \tag{4}$$

The offset between the pixel coordinate vector $p = (x, y)^T$ and the Gaussian center $\mu_i = (u_i, v_i)^T$ is denoted by $\Delta p = p - \mu_i$. The exponential component of the Gaussian response function is simplified into a scalar form to eliminate the requirement for matrix operations:

$$P(p, \mu_i, Q_i) = -\frac{1}{2}\Delta p^T Q_i \Delta p = -\frac{1}{2}[A(x - u_i)^2] \\ + 2B(x - u_i)(y - v_i) + C(y - v_i)^2. \tag{5}$$

This formulation significantly accelerates tile-based parallel computing and enables real-time rendering of tens of thousands of Gaussian primitives on GPUs.

**Differentiable Rasterization and Image Synthesis.** A differentiable α-blending process is executed to map the discrete set of Gaussian primitives back into the continuous image domain. The physical accumulation effect of light passing through multiple semi-transparent layers is simulated by this process. For a specific pixel location $p$, the set of overlapping primitives $\eta(p)$ is sorted based on depth. The final reconstructed color $C(p)$ at pixel $p$ is determined as the weighted sum of contributions from all relevant Gaussian primitives:

$$C(p) = \sum_{i \in \eta(p)} c_i \cdot \alpha_i(p) \cdot T_i(p), \tag{6}$$

where $c_i$ denotes the instantaneous RGB value of the $i$-th Gaussian primitive; and $\alpha_i(p)$ represents the instantaneous opacity at pixel, which is jointly determined by the baseline opacity and a spatial decay term:

$$\alpha_i(p) = o_i \cdot \exp\left(P(p, \mu_i, Q_i)\right). \tag{7}$$

Transmittance as light reaches the $i$-th primitive is represented by $T_i(p)$. This parameter signifies the residual light intensity not occluded by the preceding $i - 1$ primitives:

$$T_i(p) = \prod_{j=1}^{i-1}(1 - \alpha_j(p)). \tag{8}$$

The rasterization pipeline is fully differentiable. Consequently, pixel-level errors on the image plane are backpropagated to the parameters $\mu_i$, $\Sigma_i$ and $c_i$ of each Gaussian primitive via the chain rule.

**Optimization Objective and Compact Representation.** Parameter set $\Theta = \cup_i \Theta_i$ is optimized following the establishment of the 2DGS representation within the compressed image domain. The rendered image $\hat{I}$ is required to approximate the original image $I_{low}$ in both global color and structural characteristics. A composite loss function is formulated as follows:

$$\ell_{rec} = (1 - \lambda_{ssim}) \| \hat{I} - I_{low} \|_1 \\ + \lambda_{ssim}\left(1 - MS\text{-}SSIM(\hat{I}, I_{low})\right). \tag{9}$$

Gaussian primitives manifest distinct self-organizing properties during iterative optimization. These primitives automatically migrate to high-frequency image regions including object boundaries. Adjusting the covariance matrix $\Sigma_i$ facilitates the formation of elongated ellipses to fit edges accurately. A limited number of large-scale Gaussians provide sufficient coverage for smooth regions. This process ultimately yields a highly sparse parameter set $\Theta^*$.

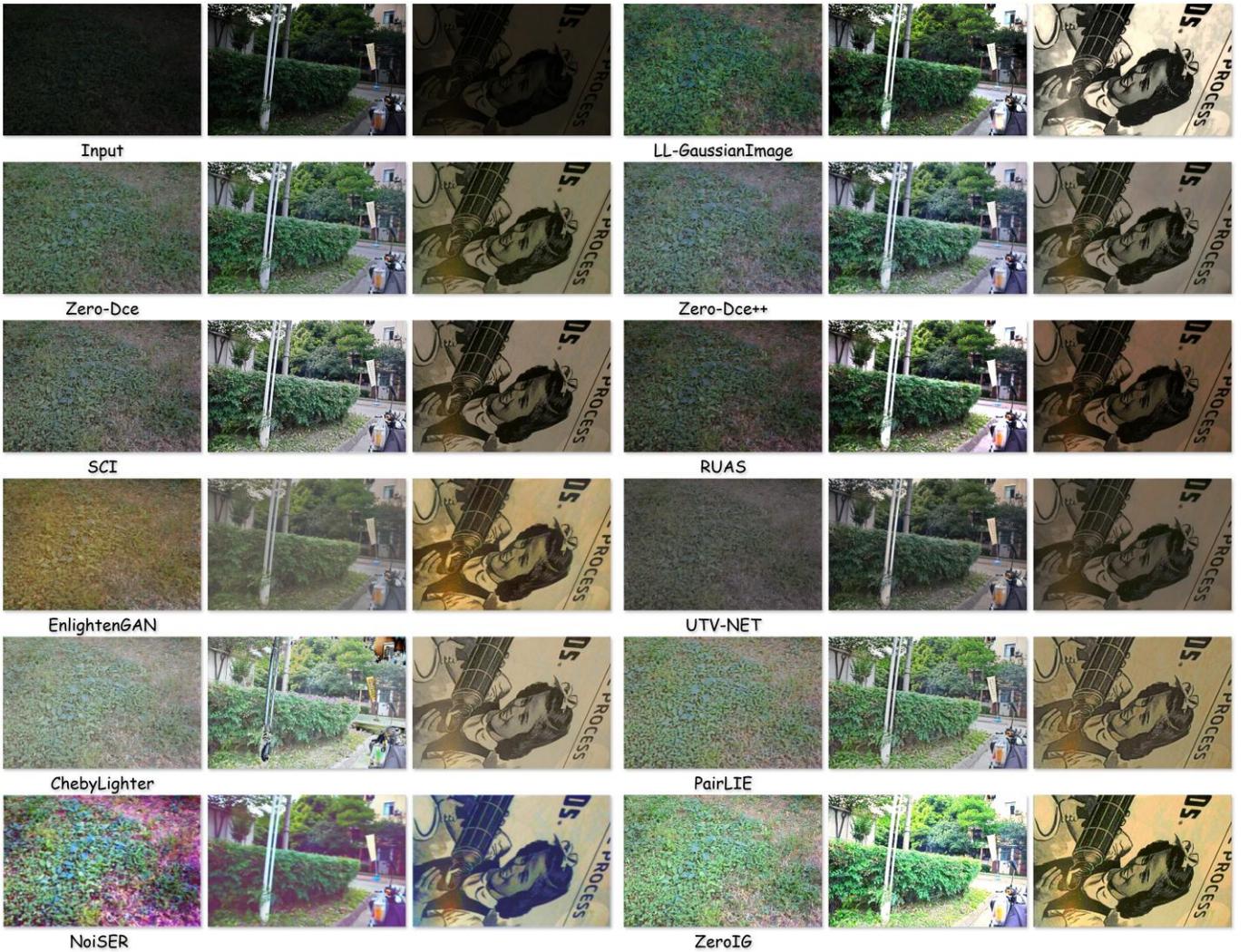

**Fig. 5.** Visual comparison of LL-GaussianImage with SOTA methods on the LOL, LSRW-HUAWEI, and LSRW-NIKON datasets.

To quantify the compactness of this explicit representation, the parameter complexity of traditional pixel grids is compared against 2D Gaussian primitives. A raw pixel representation with a resolution of $H \times W \times 3$ requires degrees of freedom totaling $DoF_{pixel} = H \times W \times 3$. By contrast, the Gaussian representation relies on the number of primitives $N$. Each primitive is characterized by nine scalar parameters: two for position, three for covariance, three for color, and one for opacity. The total degrees of freedom are thus $DoF_{gaussian} = N \times 9$. The resulting compression ratio between these representations is defined as:

$$Cr = \frac{DoF_{pixel}}{DoF_{gaussian}} = \frac{3HW}{9N} = \frac{HW}{3N}. \quad (10)$$

Approximately $7.2 \times 10^5$ numerical values are required by traditional representations for an image with a resolution of $600 \times 400$ in a typical experimental setup. By contrast, the proposed approach leverages spatial redundancy to reconstruct image structures using a minimal number of primitives. Under extremely compact settings where $N \approx 1500$, the image preserves primary structural and color characteristics. This configuration yields a compression ratio of $Cr \approx 53.3$ and achieves effective encoding of continuous image signals using less than 2% of the original parameter volume. Such significant compression capability stems from the inherent properties of Gaussian primitives. A single primitive fits large-scale and low-frequency smooth regions to eliminate the point-by-point storage of redundant information required by pixel grids. This highly decoupled and compact latent space provides the physical foundation for subsequent zero-shot illumination enhancement.

Consequently, 2DGS-based representation serves as an efficient compression format and decomposes images into a series of visual primitives with semantic potential. An ideal decoupled operation space is established for the second stage of unsupervised enhancement.

**Optimization Strategy and First-Stage Training Workflow.** The 2D image reconstruction task is performed under a cold start configuration. This approach is distinguished from 3D Gaussian Splatting which typically relies on sparse point clouds from Structure-from-Motion as geometric priors. An end-to-end adaptive training pipeline is established to facilitate convergence from an unordered random distribution to a highly structured image representation. The complete optimization process is formalized as a search problem within the parameter space $\mathfrak{R}^{9N}$.

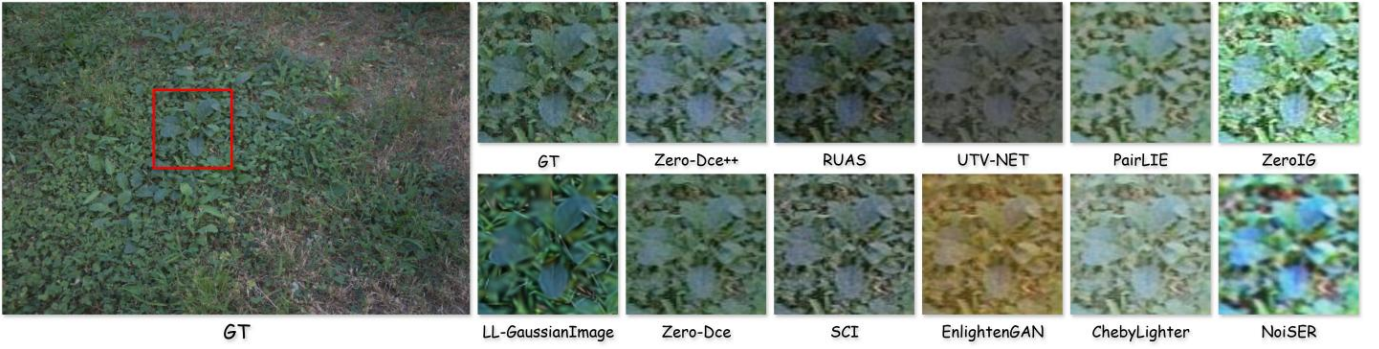

**Fig. 6.** Visual comparison of detailed features on the LOL dataset. Magnified views of the regions marked by red boxes are provided for each method.

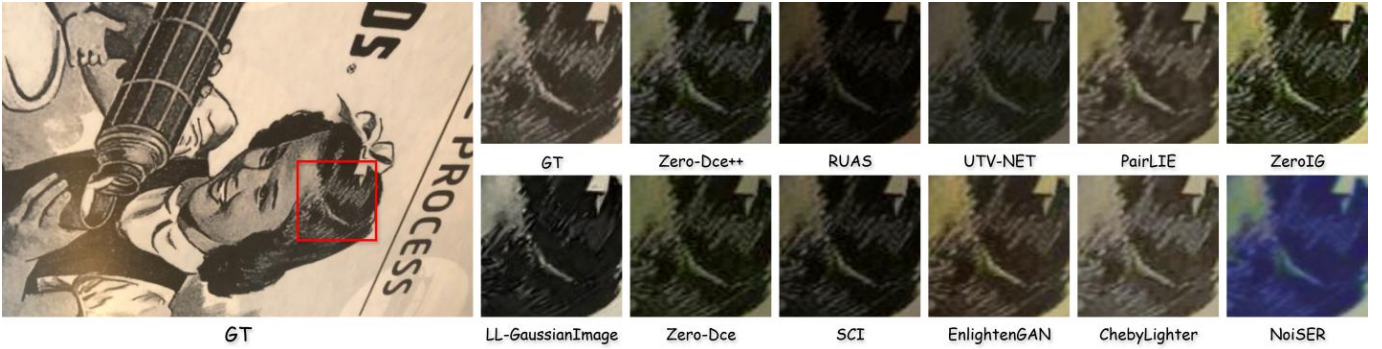

**Fig. 7.** Visual comparison of detailed features on the LSRW dataset. Magnified views of the regions marked by red boxes are provided for each method.

Prior assumptions regarding image content are excluded. Center coordinates of the Gaussian primitives are initialized as a uniform random distribution $\mu_i \sim \mu(\Omega)$ on the image plane. Covariance matric $\Sigma$ is initialized as isotropic circle while color $c$ is initialized to zero value and opacity $o$ receives a small initial value to facilitate gradient flow. This initial state compels the model to rely exclusively on illumination gradients to determine the optimal geometric layout. A forward rasterization process $f(\Theta^{(t)}) \rightarrow \hat{I}^{(t)}$ is executed in each iteration $t$ to generate the current rendered view. The reconstruction loss $\ell$ is subsequently calculated and an automatic differentiation engine computes gradients for each primitive parameter:

$$\nabla_\Theta \ell = \left[ \frac{\partial \ell}{\partial \mu_i}, \frac{\partial \ell}{\partial \Sigma_i}, \frac{\partial \ell}{\partial c_i}, \frac{\partial \ell}{\partial o_i} \right]. \tag{11}$$

Attribute parameters exhibit varying scales and sensitivities. Coordinate variations typically exert a more substantial influence on the loss function than minor color perturbations. To address these discrepancies, the Adam optimizer facilitates a joint update of all parameters. The total iteration count $T$ is set to an empirical value of 30,000.

### B. Semantically Guided Zero-Shot Unsupervised Enhancement

The discrete image primitive representation $\mathfrak{I} = \{(\mu_i, \Sigma_i, c_i)\}_{i=1}^{N}$ is acquired. Color attributes $c_i$ are optimized in the second stage to achieve image enhancement while geometric structures $\mu_i$ and $\Sigma_i$ are frozen. Three core submodules are integrated into the enhancement workflow as illustrated in Fig. 2. These components consist of a semantic-aware parameter estimation network, a set of differentiable color transformation operators and an iterative optimization strategy based on non-reference loss functions.

**Semantic-Aware Feature Extraction and Weight Estimation.** Traditional image enhancement methods typically apply a uniform transformation curve to the entire image and disregard variations in illumination response across diverse semantic regions including sky, shadows, and vegetation. Local adaptive enhancement is realized through a lightweight encoder that extracts semantic features and predicts mixing weights for each Gaussian primitive. With the low-light image $\hat{I} \in \mathfrak{R}^{H \times W \times 3}$ as input, the first seven layers of a pre-trained MobileNetV2 function as the feature extractor $E(\cdot)$ to capture low-level textures and mid-level semantic information. The feature extraction process is defined as follows:

$$F = E(I_{low}), \quad F \in \mathfrak{R}^{H' \times W' \times D}, \tag{12}$$

where $D = 32$ denotes the feature dimension; $H'$ and $W'$ signifies the resolution after downsampling. To map continuous image features onto the discrete Gaussian point cloud, a spatially aligned mixing weight head is designed. Specifically, the features $F$ are mapped by the head network $H(\cdot)$ into a weight space to generate a low-resolution weight map $M_{low} \in \mathfrak{R}^{H' \times W' \times K}$. The parameter $K$ denotes the number of predefined color transformation operators. To obtain a specific mixing weight vector $w_i \in \mathfrak{R}^K$ for each Gaussian primitive $G_i$, a bilinear interpolation operation $S(\cdot)$ is employed

TABLE I
PERFORMANCE COMPARISON RESULTS ON THE LOL DATASET, WHERE RED AND BLUE HIGHLIGHTS INDICATE SIGNIFY THE BEST AND SECOND-BEST PERFORMANCE FOR EACH INDIVIDUAL METRIC RESPECTIVELY.

| Method | SSIM↑ | PSNR↑ | LPIPS↓ | NIQE↓ | LOE↓ | DE↑ | EME↑ |
|--------|-------|-------|--------|-------|------|-----|------|
| ZeroDCE++ [15] | 0.82 | 21.68 | 0.17 | 3.83 | 13.31 | 1.75 | 10.78 |
| ZeroDCE [16] | 0.81 | 26.47 | 0.12 | 3.75 | 42.55 | 1.59 | 10.93 |
| SCI [17] | 0.81 | 20.84 | 0.18 | 4.42 | 17.91 | 1.96 | 13.49 |
| RUAS [18] | 0.81 | 20.44 | 0.14 | 3.58 | 11.06 | 1.50 | 14.34 |
| EnlightenGAN [19] | 0.81 | 18.20 | 0.30 | 3.67 | 82.64 | 1.51 | 6.17 |
| UTV-NET [22] | 0.79 | 15.22 | 0.29 | 3.44 | 21.71 | 0.97 | 6.99 |
| ChebyLighter [23] | 0.72 | 12.02 | 0.26 | 3.51 | 17.75 | 1.68 | 5.62 |
| PairLIE [25] | 0.75 | 15.74 | 0.34 | 6.49 | 72.52 | 1.72 | 7.28 |
| NoiSER [26] | 0.65 | 13.25 | 0.76 | 6.91 | 137.2 | 2.51 | 8.37 |
| ZeroIG [28] | 0.55 | 11.35 | 0.45 | 5.10 | 30.98 | 2.62 | 12.84 |
| LL-GaussImage | 0.86 | 21.71 | 0.17 | 4.66 | 12.71 | 1.52 | 13.52 |

TABLE II
PERFORMANCE COMPARISON ON THE LSRW (HUAWEI) AND LSRW (NIKON) DATASETS. RED AND BLUE INDICATE THE BEST AND SECOND-BEST RESULTS FOR EACH METRIC RESPECTIVELY.

| Method | SSIM↑ | PSNR↑ | LPIPS↓ | NIQE↓ | LOE↓ | DE↑ | EME↑ |
|--------|-------|-------|--------|-------|------|-----|------|
| ZeroDCE++ [15] | 0.74 | 11.67 | 0.15 | 3.27 | 48.84 | 1.82 | 11.52 |
| ZeroDCE [16] | 0.70 | 11.01 | 0.17 | 3.23 | 30.29 | 1.57 | 10.77 |
| SCI [17] | 0.67 | 10.96 | 0.14 | 3.48 | 12.89 | 1.73 | 13.58 |
| RUAS [18] | 0.48 | 8.32 | 0.21 | 3.83 | 10.29 | 1.45 | 15.51 |
| EnlightenGAN [19] | 0.82 | 15.56 | 0.13 | 3.21 | 55.98 | 2.02 | 5.94 |
| UTV-NET [22] | 0.64 | 9.66 | 0.17 | 3.65 | 15.83 | 1.27 | 5.93 |
| ChebyLighter [23] | 0.87 | 18.43 | 0.09 | 3.39 | 16.16 | 1.82 | 5.09 |
| PairLIE [25] | 0.82 | 14.09 | 0.14 | 4.03 | 68.58 | 1.57 | 5.87 |
| NoiSER [26] | 0.77 | 17.6 | 0.36 | 4.71 | 31.87 | 2.11 | 4.38 |
| ZeroIG [28] | 0.76 | 19.12 | 0.15 | 3.5 | 20.48 | 2.41 | 11.38 |
| LL-GaussImage | 0.88 | 18.65 | 0.12 | 3.59 | 18.72 | 1.84 | 12.08 |

to sample the weight map:

$$w_i = Soft max(S(M_{low}, \mu_i)), \qquad (13)$$

where the $Soft max(\cdot)$ operation is applied across the channel dimension to ensure that $\sum_{k=1}^{K} w_{i,k} = 1$. As illustrated in Fig. 3, the convexity of color attributes is guaranteed through this process. Concurrently, a differentiable connection between the image-space semantic context and the discrete Gaussian primitives is established via our training strategy, whereby the enhancement strategy for each Gaussian point is enabled to perceive its surrounding semantic environment.

**Residual MLP-Based Mixture-of-Experts Color Transformation.** A robust and flexible color mapping space is constructed through the learning of global color transformation operators instead of the direct regression of enhanced color values. These operators constitute a Mixture-of-Experts system where each expert is parameterized by a lightweight MLP. The $k$-th color operator is defined as $\Phi_k: \mathfrak{R}^3 \to \mathfrak{R}^3$. Each $\Phi_k$ consists of three fully connected layers and ReLU activation functions. The adoption of a residual learning mechanism ensures stability during the initial training phase and preserves the reversibility of the enhancement process. The output $c'_k$ of the $k$-th operator for an input color $c_{in}$ is defined as follows:

$$\Phi_k(c_{in}) = Clamp_{[0,1]}(c_{in} + \eta_k(c_{in})), \qquad (14)$$

where $\eta_k$ denotes the MLP network, whose final layer weights are initialized to zero to ensure that the initial transformation approximates an identity mapping; $c_i^e$ represents the enhanced color for the $i$-th Gaussian primitive, which is generated via a weighted combination of the outputs from $K - 1$ learnable operators and one identity operator according to the semantic weights $w_i$:

$$c_i^e = w_{i,id} \cdot c_i + \sum_{k=1}^{K-1} w_{i,k} \cdot \Phi_k(c_i). \qquad (15)$$

Fig. 4 illustrates the network capability to preserve original colors in specific regions through high-weight identity operators while applying complex nonlinear enhancement in other areas. Fine-grained pixel-level control is realized through this design.

**Zero-shot Unsupervised Loss Function Design.** A comprehensive unsupervised loss function $\ell_{total}$ is formulated for the enhancement stage as paired reference images are unavailable. The rasterizer projects optimized Gaussian primitives back into the image space to facilitate loss calculation within the image domain. The total loss function is defined as follows:

$$\ell_{total} = \lambda_1 \ell_{exp} + \lambda_2 \ell_{hue} + \lambda_3 \ell_{spa} + \lambda_4 \ell_{col} \\ + \lambda_5 \ell_{con} + \ell_{reg}, \qquad (16)$$

where $\ell_{exp}$ denotes the exposure loss function. Correcting the illumination level involves aligning the average image brightness with a target value $E_h$. This target value is empirically set to 0.7. $Y(I_{low})$ represents the brightness component of the image:

$$\ell_{exp} = \| E[Y(I_e)] - E_h \|_2^2. \qquad (17)$$

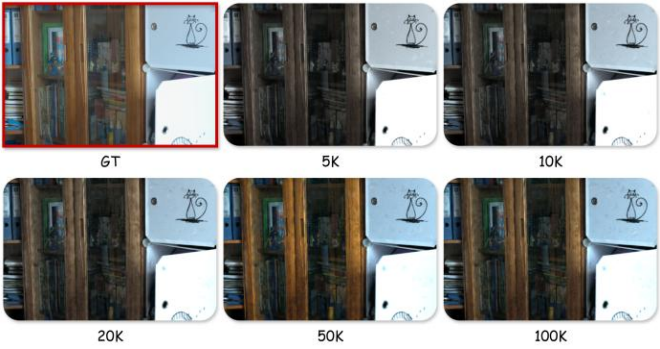

**Fig. 8.** Visual comparison of enhancement quality across different iteration counts. The red box indicates the ground truth (GT), while the others show results at 5K, 10K, 20K, 50K and 100K iterations. As optimization progresses, image details refine and brightness stabilizes. Performance saturates after 50K iterations, representing an optimal trade-off between visual quality and inference efficiency.

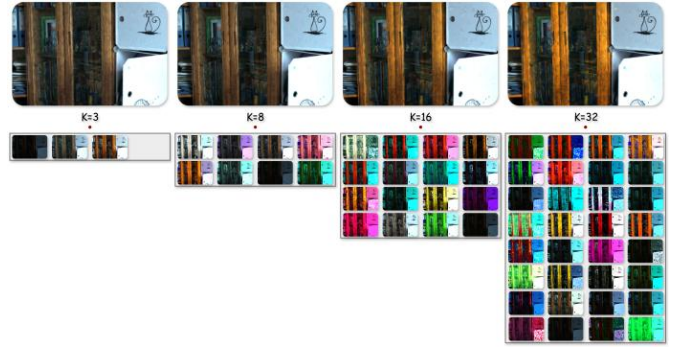

**Fig. 9.** Impact of the number of operators $K$ on image enhancement performance. The results correspond to $K \in \{3, 8, 16, 32\}$ from left to right. Image details and color fidelity improve significantly as $K$ increases and stabilize beyond $K = 16$.

A hue preservation constraint $\ell_{hue}$ based on the HSV color space is proposed to prevent color deviation during the enhancement process. Differentiable transformation from RGB to HSV is explicitly calculated as an alternative to traditional cosine similarity. A saturation mask $S_{orig}$ is incorporated to address the instability of hue computation within low-saturation regions:

$$\ell_{hue} = \frac{\sum_p l(S_{orig,p} > \tau) \cdot min(|H_{e,p} - H_{orig,p}|, l - |H_{e,p} - H_{orig,p}|)}{\sum_p l(S_{orig,p} > \tau) + \varepsilon}, \quad (18)$$

where $l(\cdot)$ denotes the indicator function; $\tau = 0.1$ represents the saturation threshold; and the $min(\Delta, 1 - \Delta)$ term is utilized to address the periodicity of the hue circle.

Severe color attenuation frequently occurs under low illumination conditions. Color richness loss $\ell_{col}$ builds on opponent color space theory to restore vivid visual effects. Maximizing the statistical dispersion of the two opponent channels enhances perceptual vividness. The formulation is defined as follows:

$$\ell_{col} = -\sqrt{\wp_{rg}^2 + \wp_{yb}^2 + \varepsilon}, \quad (19)$$

where variables $\wp_{rg}$ and $\wp_{yb}$ represent the spatial standard deviations of the $RG$ and $YB$ channels; $\varepsilon$ provides numerical stability. This constraint enables the model to effectively stretch the color distribution during illumination enhancement, thereby preventing a hazy visual appearance.

A histogram contrast loss $\ell_{con}$ is proposed for the improvement of the dynamic range. This loss term ensures that the standard deviation of the enhanced image $\sigma_e$ remains at least $\gamma$ times the original standard deviation $\sigma_{orig}$. The scaling factor $\gamma$ is set to 1.05. The formulation is defined as follows:

$$\ell_{con} = E[RELU(\gamma \cdot \sigma_{orig} - \sigma_e)]. \quad (20)$$

The contrast loss penalizes only insufficient contrast to allow further enhancement where necessary. Spatial consistency loss $\ell_{spa}$ prevents the introduction of noise or artifacts by enforcing the gradient distribution of the enhanced image to remain consistent with the original image. *Sobel* operators are utilized for the extraction of gradient maps $\nabla I$ and the calculation of $\ell_1$ loss. The formulation is defined as follows:

$$\ell_{spa} = \|\nabla_x I_e - \nabla_x I_{low}\|_1 + \|\nabla_y I_e - \nabla_y I_{low}\|_1. \quad (21)$$

Edge and texture details are effectively preserved. The regularization term $\ell_{reg}$ incorporates total variation loss $\ell_{TV}$ and entropy loss $\ell_{ent}$ to constrain the smoothness and certainty of the mixing weight map $M$. This formulation prevents truncation penalties resulting from pixel value overflow. The formulation is expressed as follows:

$$\ell_{reg} = \lambda_6 \|\nabla M\|_1 + + \lambda_7 \sum_p \sum_k \\ -w_{p,k} \, log(w_{p,k} + \varepsilon). \quad (22)$$

The training process follows a self-supervised paradigm involving iterative optimization on a single input image without external datasets. Semantic-adaptive image enhancement is realized while the geometric structure of the original scene is faithfully maintained.

## IV. Experiments

### A. Experimental Setup

**Datasets.** The LOL and LSRW datasets [53-54] facilitate the experimental evaluation of LL-GaussianImage. The LOL dataset serves as the first paired collection for supervised low-light illumination enhancement and integrates synthetic and real-world imagery. The LSRW dataset represents the first large-scale real-world paired repository. This collection consists of two distinct subsets captured using HUAWEI P40 Pro and NIKON D7500 equipment.

**Implementation Details.** The overall network architecture of LL-GaussianImage is implemented within the PyTorch framework and executed on a single NVIDIA RTX 3090 GPU for both training and inference. Parameter settings and training strategies for the two-stage process are detailed to ensure experimental reproducibility [55]. The methodology employs a two-stage optimization paradigm to perform zero-shot instance-level optimization for each input image.

The first stage involves the initialization of $N = 70,000$ 2D Gaussian points. Optimization is conducted via the Adam optimizer with an initial learning rate of 0.01. A StepLR scheduler manages the learning rate by applying a decay factor of 0.9 every 7,000 iterations. This reconstruction process spans 30,000 iterations in total. The loss function $\ell$ follows the configuration defined in LIG [1].

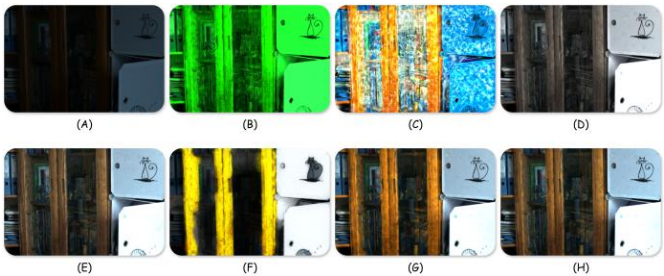

**Fig. 10.** Visual comparison of different loss functions on enhancement results. (A)–(G) show the enhancement results when removing $\ell_{exp}$, $\ell_{hue}$, $\ell_{spa}$, $\ell_{col}$, $\ell_{con}$, $\ell_{TV}$ and $\ell_{ent}$ respectively. (H) displays the result produced by LL-GaussianImage.

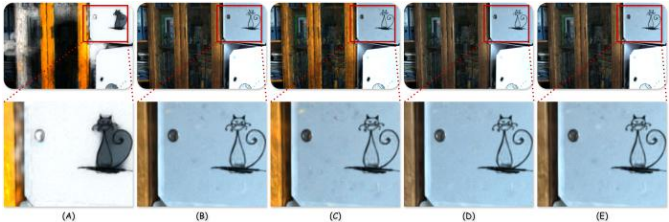

**Fig. 11.** Visualization of different weights for $\ell_{TV}$. (A)–(E) illustrate the results for $\lambda_6$ values of 0, 100, 200, 300 and 500 respectively. The bottom row displays magnified views corresponding to each value to highlight local structural details.

The second stage freezes the Gaussian geometric parameters $\mu_i$ and $\Sigma_i$ from the initial stage and adjusts appearance color only by optimizing the unified enhancement framework to adjust appearance color. The first seven layers of a pre-trained MobileNetV2 are utilized as the feature extraction backbone. Feature dimensions are mapped to $D = 32$. Illumination enhancement involves $K = 16$ MLP-based residual color operators. This stage executes 50,000 iterations using the Adam optimizer. An initial learning rate of 0.002 decays to 1% of the starting value through a cosine annealing strategy. Loss function weight coefficients are defined as $\lambda_1 = 50$, $\lambda_2 = 25$, $\lambda_3 = 32.5$, $\lambda_4 = 2.91$, $\lambda_5 = 0.44$, $\lambda_6 = 300$, $\lambda_7 = 0.01$. Specific adjustments depend on the illumination levels of the low-light images. 2D Gaussian rendering is performed via a tile-based rasterizer with a tile size of $16 \times 16$. Reflection padding is applied to the image boundaries during feature extraction and gradient computation to eliminate edge artifacts.

Evaluation Metrics. Three widely recognized full-reference image metrics and four non-reference image metrics are selected within the field of image restoration and enhancement. Consistency between enhanced images and reference images for datasets with paired ground truth is measured through three full-reference metrics. These consist of Peak Signal-to-Noise Ratio (PSNR), Structural Similarity (SSIM) and Learned Perceptual Image Patch Similarity (LPIPS). PSNR quantifies pixel-level fidelity. SSIM evaluates image similarity based on luminance, contrast, and structure. LPIPS measures the distance between enhanced and reference images within the feature space of a VGG network.

Four non-reference image metrics are employed to evaluate image naturalness, contrast and information content, particularly for performance in unpaired data or real-world scenarios. These metrics consist of Natural Image Quality Evaluator (NIQE), Lightness Order Error (LOE), Discrete Entropy (DE) and Enhancement Measure Evaluation (EME). The degree of image naturalness is quantified by NIQE through the measurement of the distance between enhanced images and a statistical model of natural images. LOE evaluates naturalness preservation during the illumination enhancement process. The richness of information contained within the images is measured by DE. EME is defined as a non-reference metric based on Weber's law to quantify local contrast variations.

### B. Performance Comparison

Ten SOTA unsupervised methods within the LLIE domain are selected for comparison with LL-GaussianImage to ensure fairness [15-19, 22-23, 25-26, 28]. Evaluation results for the LOL and LSRW datasets are reported in TABLE I and TABLE II [53-54].

Visual comparison results for images selected from the LOL, LSRW (HUAWEI) and LSRW (NIKON) datasets are illustrated in Fig. 5. LL-GaussianImage achieves significant illumination enhancement while avoiding the underexposure observed in RUAS and UTV-Net or the overexposure found in ZeroIG. Regarding clarity, SCI, LL-GaussianImage, and ChebyLighter demonstrate superior visual quality. Conversely, EnlightenGAN produces haze-like artifacts. In terms of color reproduction, LL-GaussianImage exhibits slight oversaturation but remains free from the severe color bias present in NoiSER and EnlightenGAN. These visual results establish LL-GaussianImage as a robust framework for illumination enhancement within the compressed image domain.

The advantages of LL-GaussianImage regarding detail restoration are further validated through complex scenes from the LOL and LSRW datasets illustrated in Fig. 6 and Fig. 7. Leaf texture preservation by LL-GaussianImage is slightly inferior to pixel-domain low-light enhancement methods in Fig. 6. However, LL-GaussianImage maintains high similarity to ground-truth references in color saturation and contrast. Outstanding detail restoration is also exhibited by LL-GaussianImage in Fig. 7. Such performance is noteworthy because the data volume of compressed images is dozens of times smaller than that of pixel-domain images.

The evaluation framework incorporates full-reference metrics including PSNR, SSIM, and LPIPS, along with no-reference metrics such as NIQE, LOE, DE, and EME. TABLE I and TABLE II summarize the quantitative performance across four datasets. LL-GaussianImage consistently ranks among the top two performers in terms of PSNR and SSIM. Superior performance is also demonstrated by the proposed method across the rankings of other evaluation metrics.

LL-GaussianImage balances reconstruction quality and compression performance despite certain limitations in detail preservation for specific scenes. This framework achieves a maximum compression ratio of 50x and represents the inaugural solution for low-light illumination enhancement within the 2DGS compressed domain. Consequently, this approach maintains a competitive advantage over various SOTA methods through its unique integration of efficiency and efficacy.

TABLE III
Performance evaluation of LL-GaussianImage under different iteration counts. Red and blue indicate the best and second-best results for each metric respectively.

| Enhance Iterations | SSIM↑ | PSNR↑ | LPIPS↓ | NIQE↓ | LOE↓ | DE↑ | EME↑ |
|---|---|---|---|---|---|---|---|
| 5K | 0.74 | 17.58 | 0.24 | 3.61 | 45.03 | 0.11 | 8.07 |
| 10K | 0.73 | 17.55 | 0.24 | 3.53 | 56.85 | 0.12 | 7.27 |
| 20K | 0.73 | 18.48 | 0.23 | 3.48 | 78.66 | 0.1 | 7.24 |
| 50K | 0.76 | 18.56 | 0.21 | 3.44 | 68.41 | 0.25 | 7.26 |
| 100K | 0.75 | 18.52 | 0.24 | 3.46 | 66.62 | 0.16 | 7.25 |

TABLE IV
Performance evaluation of LL-GaussianImage under different iteration counts. Red and blue indicate the best and second-best results for each metric respectively.

| $K$ | SSIM↑ | PSNR↑ | LPIPS↓ | NIQE↓ | LOE↓ | DE↑ | EME↑ |
|---|---|---|---|---|---|---|---|
| 3 | 0.71 | 16.4 | 0.25 | 3.48 | 74.96 | 0.14 | 7.14 |
| 8 | 0.73 | 18.42 | 0.29 | 3.49 | 82.23 | 0.16 | 7.12 |
| 16 | 0.76 | 18.56 | 0.21 | 3.44 | 68.41 | 0.25 | 7.26 |
| 32 | 0.74 | 17.43 | 0.27 | 3.45 | 64.33 | 0.14 | 7.03 |

TABLE V
Evaluation results for different weight values of $\ell_{TV}$. Red and blue indicate the best and second-best results for each metric respectively.

| $\lambda_6$ | SSIM↑ | PSNR↑ | LPIPS↓ | NIQE↓ |
|---|---|---|---|---|
| 0 | 0.49 | 11.76 | 0.42 | 4.00 |
| 100 | 0.74 | 18.46 | 0.23 | 3.48 |
| 200 | 0.70 | 17.68 | 0.25 | 3.45 |
| 300 | 0.75 | 18.60 | 0.21 | 3.58 |
| 500 | 0.76 | 18.63 | 0.20 | 3.46 |

### C. Ablation Study

A series of extensive ablation experiments was conducted on the LOL dataset to validate the effectiveness of individual components and the rationality of hyperparameter settings within the LL-GaussianImage framework. The baseline setup incorporates 16 color operators and 30,000 training iterations while utilizing the full set of proposed loss functions.

**Impact of Enhancement Iterations.** Optimal illumination enhancement parameters are fitted through iterative optimization on individual images due to the zero-shot learning strategy. Convergence and visual quality are directly determined by the iteration count. Model performance was evaluated across settings of 5,000, 10,000, 20,000, 50,000 and 100,000 iterations. Fig. 8 illustrates that brightness recovery and detail clarity improve as the iteration count increases, peaking at 50K before exhibiting a slight decline. Consequently, 50K iterations provide optimal brightness and color fidelity. Furthermore, quantitative results in TABLE III demonstrate that the 50K configuration yields the superior performance across most metrics, with the exceptions of LOE and EME. Based on these observations, the second stage adopts 50,000 iterations as the standard setting.

**Effectiveness of Operator Quantity.** The final enhanced image is synthesized through the blending of outputs from a Mixture-of-Experts system using spatially adaptive weights. Fitting capacity for complex illumination mappings is determined by the operator quantity $K$. Model performance is evaluated across $K \in \{3, 8, 16, 32\}$ to identify the optimal configuration. Image details and colors are improved as $K$ increases according to Fig. 9. These attributes stabilize after $K = 16$ while occasionally exhibiting signs of oversaturation. As shown in TABLE IV, the configuration with $K = 16$ achieves the best results across all metrics except for LOE. Consequently, $K = 16$ is selected as the optimal value for the proposed framework.

**Importance of Loss Functions.** Multiple components for exposure, color, structure and regularization are incorporated into the loss function. Systematic removal of each term demonstrates its specific contribution to the enhancement process. Visualized results are summarized in Fig. 10. Each loss component is critical for generating high-quality images. Omitting the exposure loss $\ell_{exp}$ prevents effective illumination recovery. The removal of the hue preservation loss $\ell_{hue}$, spatial consistency loss $\ell_{spa}$ and contrast loss $\ell_{con}$ leads to significant color shifts and poor contrast stability. Furthermore, the absence of the total variation loss $\ell_{TV}$ hinders smooth color transitions between Gaussian primitives. Therefore, all loss functions are consequently utilized in this framework.

**Analysis of Specific Regularization Terms.** Sensitivity to the weight hyperparameter of $\ell_{TV}$ is specifically examined in addition to the general losses mentioned previously. $\ell_{TV}$ constrains spatial smoothness of the mixture weight maps extracted by MobileNet. Color transitions of the ellipsoids become more uniform as the value of $\lambda_6$ increases and reach stability at 300 as depicted in Fig. 11.

Magnified views indicate that smaller values of $\lambda_6$ lead to disorganized primitive distributions and local illumination artifacts. Consequently, $\lambda_6 = 500$ serves as the final weight for $\ell_{TV}$. Optimal performance is achieved with $\lambda_6 = 500$ across all metrics except NIQE as shown in TABLE V.

### D. Limitation and Future works

Significant potential and generalizability are exhibited by LL-GaussianImage. However, optimization for rapid inference alongside image quality preservation is necessitated by the advancement of feed-forward Gaussian Splatting. Furthermore, the optimization of Gaussian ellipsoids for detailed texture reconstruction remains an unresolved challenge. LL-GaussianImage is expected to facilitate advancements in low-level vision applications within the 2DGS compressed image domain.

## V. CONCLUSION

LL-GaussianImage is formulated as the first zero-shot unsupervised framework for low-light illumination enhancement performed directly within the 2DGS compressed domain. The introduction of a semantic-guided Mixture-of-Experts module and a decoupling strategy for geometry and appearance transforms the enhancement task into manifold mapping within the attribute space. This framework achieves high-fidelity compression-as-enhancement while maintaining a frozen geometric structure. Such a paradigm avoids the computational redundancy and secondary distortion inherent in the traditional decompression-enhancement-recompression workflow. To address speed bottlenecks in iterative optimization and preserve fine-grained features, future work will focus on developing feed-forward Gaussian Splatting architectures, thereby establishing novel pathways for low-level vision tasks within the 2DGS compressed domain.

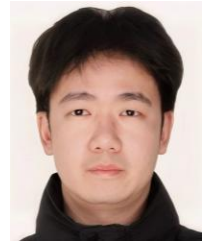

**Yuhan Chen** received his master's degree in 2024 from the College of Mechanical Engineering at Chongqing University of Technology. He is currently pursuing the Ph.D. degree in College of Mechanical and Vehicle Engineering at Chongqing University, China. His research interests include deep learning, Low-level Vision and Gaussian Splatting.

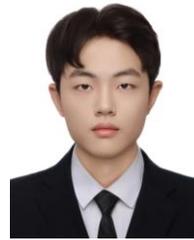

**Wenxuan Yu** received the B.E. degree majoring in Mechanical Design, Manufacturing, and Automation at Chongqing University in 2025. He is currently pursuing the M.E. degree in Mechanical Engineering at Chongqing University, Chongqing, China. His research interests include computer vision, Gaussian Splatting and deep learning.

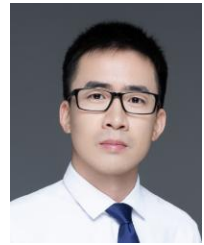

**Guofa Li** received the Ph.D. degree in Mechanical Engineering from Tsinghua University, China, in 2016. He is currently a Professor with Chongqing University, China. His research interests include environment perception, driver behavior analysis, and smart decision-making based on artificial intelligence technologies in autonomous vehicles and intelligent transportation systems. He serves as the Associate Editor for *IEEE Transactions on Intelligent Transportation Systems, IEEE Transactions on Affective Computing,* and *IEEE Sensors Journal.*

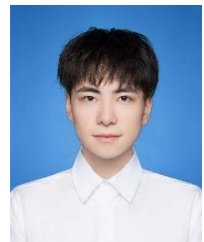

**Yijun Xu** received the B.E. degree from Southwest University, Chongqing, China, in 2024. He is currently pursuing the M.S. degree in Electronic Information at the School of Electronic Information, Wuhan University, Wuhan, China. His research interests include machine vision and signal processing.

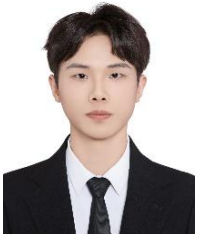

**Ying Fang** received the B.E. degree majoring in Vehicle Engineering at Chongqing University of Technology. He is currently pursuing the M.E. degree in Mechanical Engineering at Chongqing University, Chongqing, China. His research interests include computer vision, Gaussian Splatting and deep learning.

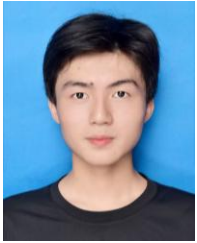

**Yicui Shi** received the B.E. degree majoring in Automotive Engineering at Chongqing University in 2025. He is currently pursuing the M.E. degree in Automotive Engineering at Chongqing University, Chongqing, China. His research interests include computer vision, Gaussian Splatting and deep learning.

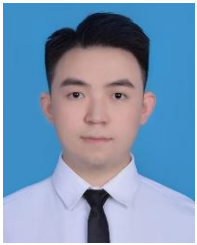

**Long Cao** received his master's degree from the School of Mechanical Engineering, Guangxi University, in 2025. He is currently pursuing the Ph.D. degree at the College of Mechanical and Vehicle Engineering, Chongqing University. His research interests include computer graphics, inverse rendering, and deep learning.

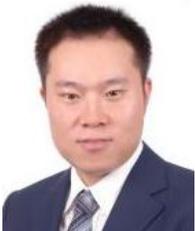

**Wenbo Chu** received his B.S. degree majored in Automotive Engineering from Tsinghua University, China, in 2008, and his M.S. degree majored in Automotive Engineering from RWTH-Aachen, German and Ph.D. degree majored in Mechanical Engineering from Tsinghua University, China, in 2014.

He is currently a research fellow at Western China Science City Innovation Center of Intelligent and Connected Vehicles (Chongqing) Co, Ltd., and National Innovation Center of Intelligent and Connected Vehicles.

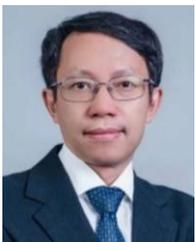

**Keqiang Li** received the B.E. degree from Tsinghua University, Beijing, China, in 1985, and the M.E. and Ph.D. degrees from Chongqing University, Chongqing, China, in 1988 and 1995, respectively.

He is currently a Professor with the School of Vehicle and Mobility, Tsinghua University. He is the Chief Scientist of Intelligent and Connected Vehicle Innovation Center of China, and the Director of State Key Laboratory of Automotive Safety and Energy of China. His current research interests include intelligent connected vehicles, cloud-based control for vehicles, and vehicle dynamics systems.